\documentclass[10pt,twocolumn,letterpaper]{article}

\usepackage{cvpr}
\usepackage{times}
\usepackage{epsfig}
\usepackage{graphicx}
\usepackage{amsmath}
\usepackage{amssymb}
\usepackage{subfig}
\usepackage{xspace}
\usepackage{bbold}
\usepackage{xcolor}
\usepackage{multirow}
\usepackage[numbers,sort]{natbib}

\newcommand{\vct}[1]{\ensuremath{\boldsymbol{#1}}}

\newcommand{\set}[1]{\ensuremath{\mathcal{#1}}}
\newcommand{\con}[1]{\ensuremath{\mathsf{#1}}}

\newcommand{\argmax}{\operatornamewithlimits{\arg\,\max}}

\DeclareMathOperator{\Var}{Var}
\DeclareMathOperator{\E}{E}
\DeclareMathOperator{\DA}{DA}
\DeclareMathOperator{\mDA}{mDA}

\usepackage[pagebackref=true,breaklinks=true,letterpaper=true,colorlinks,bookmarks=false]{hyperref}

\cvprfinalcopy 

\def\cvprPaperID{3548} 

\ifcvprfinal\fi
\begin{document}

\title{
Boosting Domain Adaptation by Discovering Latent Domains
}

\author{Massimiliano Mancini$^{1,2}$, Lorenzo Porzi$^3$, Samuel Rota Bul\`o$^3$, Barbara Caputo$^{1,4}$, Elisa Ricci$^{2,5}$\\
$^1$Sapienza University of Rome, $^2$Fondazione Bruno Kessler, $^3$Mapillary Research,\\$^4$Italian Institute of Technology, $^5$University of Trento\\
{\tt\small \{mancini,caputo\}@diag.uniroma1.it}, {\tt\small\{lorenzo,samuel\}@mapillary.com}, {\tt\small eliricci@fbk.eu}
}

\newcommand{\myparagraph}[1]{\vspace{5pt}\noindent\textbf{#1}}

\maketitle

\begin{abstract}
Current Domain Adaptation (DA) methods based on deep architectures assume that the source samples arise from a single distribution. However, in practice most datasets can be regarded as mixtures of multiple domains. In these cases exploiting single-source DA methods for learning target classifiers may lead to sub-optimal, if not poor, results. In addition, in many applications it is difficult to manually provide the domain labels for all source data points, \ie latent domains should be automatically discovered.
This paper introduces a novel Convolutional Neural Network (CNN) architecture which (i) automatically discovers latent domains in visual datasets and (ii) exploits this information to learn robust target classifiers. 
Our approach is based on the introduction of two main components, which can be embedded into any existing CNN architecture: (i) a side branch that automatically computes the assignment of a source sample to a latent domain and (ii) novel layers that exploit domain membership information to appropriately align the distribution of the CNN internal feature representations to a reference distribution.
We test our approach on publicly-available datasets, showing that it outperforms state-of-the-art multi-source DA methods by a large margin.
\end{abstract}

\vspace{-30pt}
\section{Introduction}
The problem that trained models perform poorly when tested on data from a different distribution is commonly referred to as \textit{domain shift}. This issue is especially relevant in computer vision, as visual data is characterized by large appearance variability, \eg due to differences in resolution, changes in camera pose, occlusions and illumination variations. To address this problem, several transfer learning and domain adaptation approaches have been proposed in the last decade \cite{pan2010survey}. 

Domain Adaptation (DA) methods are specifically designed to transfer knowledge from a \textit{source} domain to the domain of interest, \ie the \textit{target} domain, in the form of learned models or invariant feature representations. The problem has been widely studied and both theoretical results \cite{ben2010theory,mansour2009domain} and several shallow  \cite{huang2006correcting,gong2013connecting,gong2012geodesic,long2013transfer,fernando2013unsupervised} and deep learning algorithms have been developed \cite{long2015learning,tzeng2015simultaneous,ganin2014unsupervised,long2016unsupervised,ghifary2016deep,carlucci2017autodial,bousmalis2016domain}. While deep neural networks tend to produce more transferable and domain-invariant features, previous works \cite{donahue2014decaf} have shown that the domain shift is only alleviated but not removed.

\begin{figure}[t]
  \centering
  \includegraphics[width=\columnwidth,trim={0cm 0cm 0cm 0cm},clip]{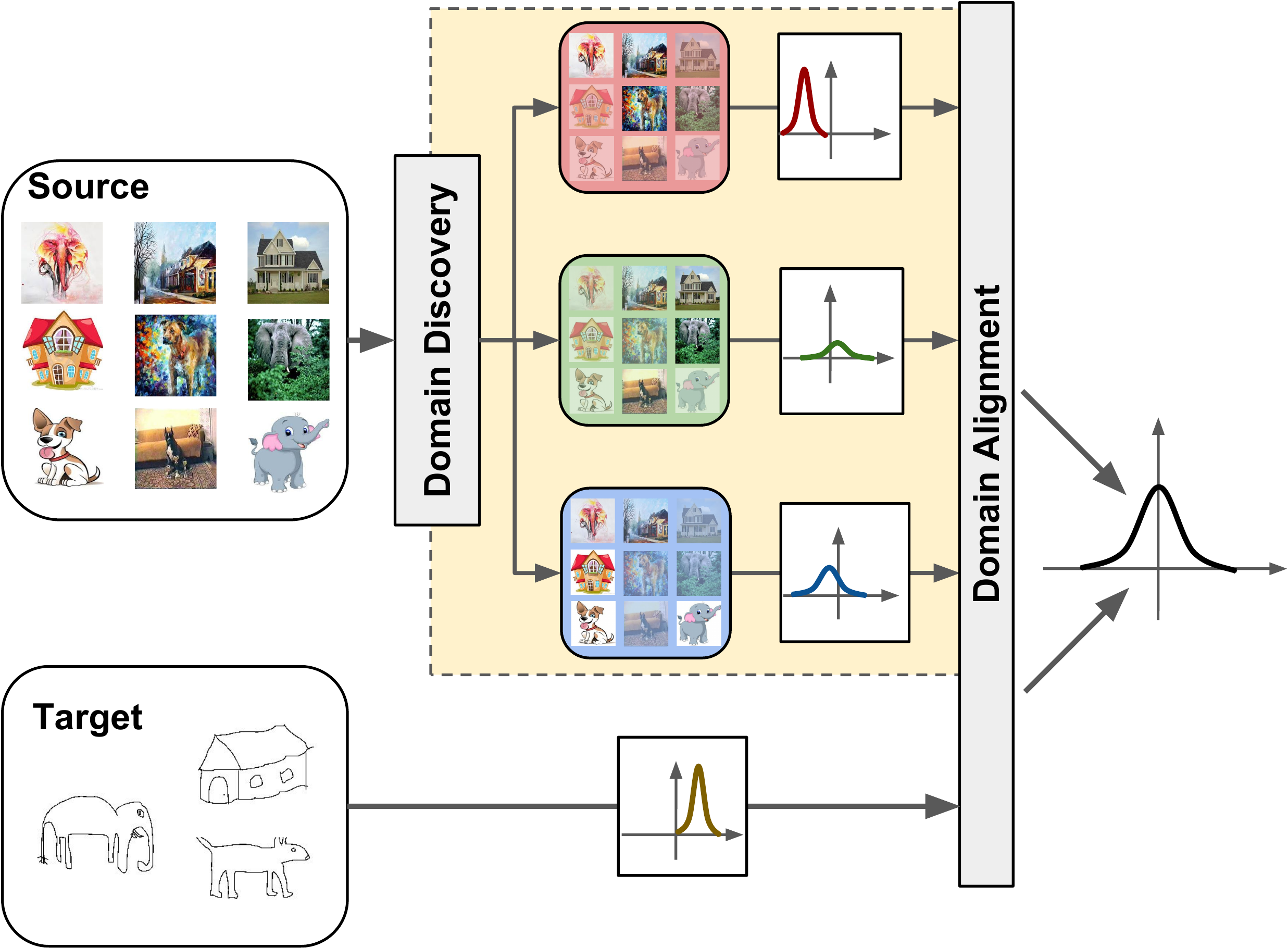}
  \caption{The idea behind our framework. 
We propose a novel deep architecture which, given a set of images, automatically discover multiple latent source domains
and use this information to align the distributions of the internal CNN feature representations of sources and target domains for the purpose of domain adaptation. Image better seen at magnification.
  }
  \label{fig:teaser}
  \vspace{-10pt}
\end{figure}

Most works on DA focus on a single-source and single-target scenario. However, in many computer vision applications 
labeled training data is often generated from multiple distributions, \ie there are multiple source domains. Examples of multi-source DA problems arise when the source set corresponds to images taken with different cameras, collected from the web or associated to multiple points of views. In these cases, a naive application of single-source DA algorithms would not suffice, leading to poor results. Therefore, in the past several research efforts have been devoted to develop DA methods operating on multiple sources \cite{mansour2009domain,duan2009domain,sun2011two}. These approaches assume that the different source domains are known. A more challenging problem arises when labeled training data correspond to latent source domains, \ie we can make a reasonable estimate on the number of source domains available, but we have no information, or only partial,  about domain labels.
To address this problem, known in the literature as \emph{latent domain discovery}, previous works have proposed methods which simultaneously discover hidden source domains and use them to learn the target classification models \cite{hoffman2012discovering,gong2013reshaping,xiong2014latent}.

This paper introduces the first deep approach able to automatically discover latent source domains in multi-source domain adaptation settings.
Our method is inspired by the recent works~\cite{carlucci2017autodial,carlucci2017just}, which revisit Batch Normalization layers \cite{ioffe2015batch} for the purpose of domain adaptation, introducing specific Domain Alignment layers (DA-layers). 
The main idea behind DA-layers is to cope with domain shift by aligning representations of source and target distributions to a reference Gaussian distribution. Our approach develops from the same intuition. However, to address the additional challenges of discovering and handling multiple latent domains, we propose a novel architecture which is able to (i) learn a set of assignment variables which associate source samples to a latent domain and (ii) exploit this information for aligning the distributions of the internal CNN feature representations and learn a robust target classifier (Fig.\ref{fig:teaser}).
Our experimental evaluation shows that the proposed approach alleviates the domain discrepancy and outperforms previous multi-source DA techniques on popular benchmarks, such as Office-31~\cite{saenko2010adapting} and Office-Caltech~\cite{gong2012geodesic}.


\section{Related Work}
\label{sec:related}
\vspace{-5pt}
\myparagraph{DA methods with hand-crafted features.} Earlier DA approaches operate on hand-crafted features and attempt to reduce  
the discrepancy between the source and the target domains by adopting different strategies. For instance, instance-based methods \cite{huang2006correcting,yamada2012no,gong2013connecting} develop from the idea of learning classification/regression models by re-weighting source samples according to their similarity with the target data. A different strategy is exploited by feature-based methods, coping with domain shift by learning a common subspace for source and target data such as to obtain domain-invariant representations \cite{gong2012geodesic,long2013transfer,fernando2013unsupervised}. Parameter-based methods \cite{yang2007adapting} address the domain shift problem by discovering a set of shared weights between the source and the target models. However, they usually require labeled target data which is not always available.

While most earlier DA approaches focus on a single-source and single-target setting, some works have considered the related problem of learning classification models when the training data spans multiple domains \cite{mansour2009domain,duan2009domain,sun2011two}. The common idea behind these methods is that when source data arises from multiple distributions, adopting a single source classifier is suboptimal and improved performance can be obtained leveraging information about multiple domains. However, these methods assume that the domain labels for all source samples are known in advance. In practice, in many applications the information about domains is hidden and latent domains must be discovered into the large training set.
Few works have considered this problem in the literature. Hoffman \etal \cite{hoffman2012discovering} address this task by modeling domains as  Gaussian distributions in the feature space and by estimating the membership of each training sample to a source domain using an iterative approach. 
Gong \etal \cite{gong2013reshaping} discover latent domains by devising a nonparametric approach which aims at
simultaneously achieving maximum distinctiveness among domains 
and ensuring that strong discriminative models are learned for each latent domain. In \cite{xiong2014latent} 
domains are modeled as manifolds and source images representations are learned decoupling information about semantic category and domain. By exploiting these representations the domain assignment labels are inferred using a mutual information based clustering method. 


\myparagraph{Deep Domain Adaptation.} Most recent works on DA consider deep architectures and robust domain-invariant features are learned using either supervised neural networks \cite{long2015learning,tzeng2015simultaneous,ganin2014unsupervised,ghifary2016deep,bousmalis2016domain,carlucci2017autodial}, deep autoencoders \cite{zeng2014deep} or generative adversarial networks \cite{bousmalis2016unsupervised,shrivastava2017learning}. For instance, some methods attempt to align source and target features by minimizing Maximum Mean Discrepancy  \cite{long2015learning,long2016unsupervised,sun2016deep}. Other approaches operate in a domain-adversarial setting, \ie learn domain-agnostic representations by maximizing a domain confusion loss \cite{tzeng2015simultaneous,ganin2014unsupervised}. 
Domain separation networks are proposed in \cite{bousmalis2016domain}, where feature representations are learned by decoupling the domain-specific component from a shared one.
DA-layers are described in \cite{carlucci2017just} which, embedded into an arbitrary CNN architecture, are able to align source and target representation distributions. 

While recent deep DA methods significantly outperform approaches based on hand-crafted features, they only consider single-source, single-target settings. To our knowledge, this is the first work proposing a deep architecture for discovering latent source domains and exploiting them for improving classification performance on target data.



\section{Method}
\label{sec:method}

\subsection{Problem Formulation and Notation}
In this paper we are interested in predicting labels from an output space $\set{Y}$ (\emph{e.g.} object or scene categories), given elements of an input space $\set{X}$ (\emph{e.g.} images).
We further assume that our data belongs to one of several domains: the $\con{k}$ \emph{source} domains, characterized by unknown probability distributions $p_{\mathtt{xy}}^{s_1},\dots,p_{\mathtt{xy}}^{s_\con{k}}$ defined over $\set{X}\times\set{Y}$, and the \emph{target} domain, characterized by $p_{\mathtt{xy}}^t$.
Note that the number of source domains $\con{k}$ is not necessarily known a-priori, and is left as an hyper-parameter of our method.
During training we are given a set of labeled samples from the source domains, and a set of unlabeled samples from the target domain, while we have partial or no access to domain assignment information for the source samples.
More formally, we model the source data as a set $\set{S}=\{(x_1^s,y_1^s),\dots,(x_\con{n}^s,y_\con{n}^s)\}$ of i.i.d. observations from a mixture distribution $p_{\mathtt{xy}}^s=\sum_{i=1}^\con{k} \pi_{s_i} p_{\mathtt{xy}}^{s_i}$, where $\pi_{s_i}$ is the probability of sampling from a source domain $s_i$.
Similarly, the target samples $\set{T}=\{x_1^t,\dots,x_\con{m}^t\}$ are i.i.d. observations from the marginal $p_\mathtt{x}^t$.
Furthermore, we denote by $x_\set{S}=\{x_1^s,\dots,x_\con{n}^s\}$ and $y_\set{S}=\{y_1^s,\dots,y_\con{n}^s\}$, the source data and label sets, respectively.
We assume to know the domain label for a (possibly empty) subset $\hat {\set S}\subset \set S$ of source data samples and we denote by $d_{\hat {\set S}}$ the domain labels in $\{s_1.\ldots,s_\con k\}$ of the sample points in $x_{\hat S}$. The set of domains labels, including target domain, is given by $\set D=\{s_1,\ldots,s_\con k,t\}$.

Our main goal is to learn a predictor that is able to classify samples from the target domain.
When tackling this problem we have to deal with three main difficulties: (i) the distributions of source(s) and target can be drastically different, making it hard to apply a classifier learned on one domain to the others, (ii) we lack direct observation of target labels, and (iii) the assignment of each source sample to its domain is unknown, or known for a very limited number of samples only.

Several previous works~\cite{long2015learning,tzeng2015simultaneous,ganin2014unsupervised,ghifary2016deep,bousmalis2016domain,carlucci2017autodial} have tackled the related problem of domain adaptation in the context of deep neural networks, dealing with (i) and (ii) in the case in which all source data comes from a single domain.
In particular, some recent works have demonstrated a simple yet effective approach based on the replacement of standard Batch Normalization layers with specific \textit{Domain Alignment layers} ~\cite{carlucci2017just,carlucci2017autodial}.
These layers aim to reduce internal domain shift at different levels within the network by re-normalizing features in a domain-dependent way, matching their distributions to a pre-determined one.
In the following sections we show how the same idea can be revisited to naturally tackle the case of multiple, unknown source domains.
In particular, we propose a novel Multi-domain DA layer (mDA-layer), detailed in Section~\ref{sec:dalayers}, which is able to re-normalize the multi-modal feature distributions encountered in our setting.
To do this, our mDA-layers exploit a side-output branch we attach to the main network (see Section~\ref{sec:domain-prediction}), which predicts domain assignment probabilities for each input sample.
Finally, in Section~\ref{sec:loss} we show how the predicted domain probabilities can be exploited, together with the unlabeled target samples, to construct a prior distribution over the network's parameters which is then used to define the training objective for our network.

\subsection{Multi-domain DA-layers}
\label{sec:dalayers}

\begin{figure*}[t]
  \centering
  \includegraphics[width=\textwidth]{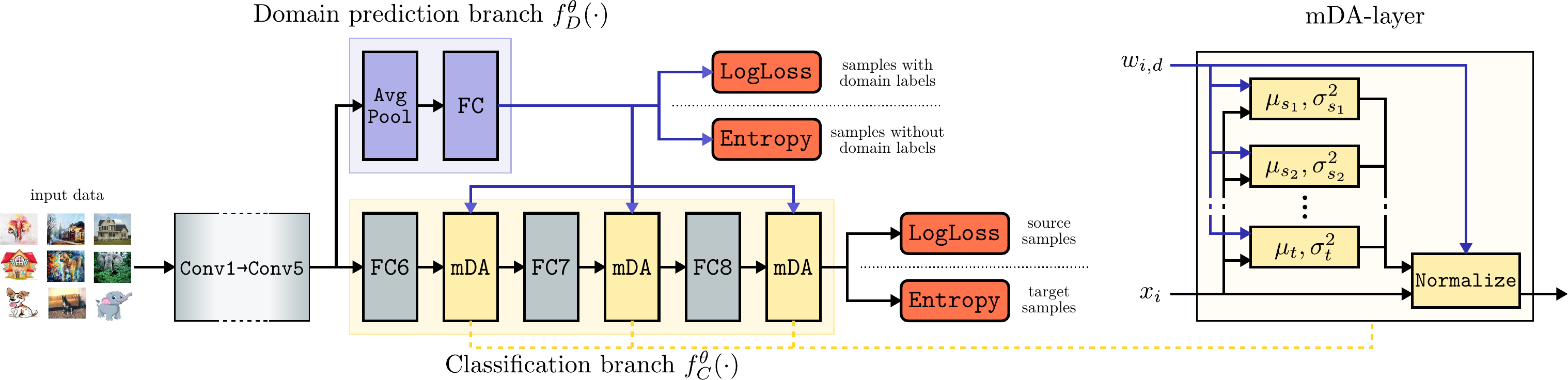}
  \caption{Schematic representation of our method applied to the AlexNet architecture (left) and of an mDA-layer (right).}
  \label{fig:teaser}
  \vspace{-10pt}
\end{figure*}

DA-layers~\cite{li2016revisiting,carlucci2017just,carlucci2017autodial} are motivated by the observation that, in general, activations within a neural network follow domain-dependent distributions.
As a way to reduce domain shift, the activations are thus normalized in a domain-specific way, shifting them according to a parameterized transformation in order to match their first and second order moments to those of a reference distribution, which is generally chosen to be normal with zero mean and unit standard deviation.
While previous works only considered settings with two domains, \ie source and target, the basic idea can in fact be applied to any number of domains, as long as the domain membership of each sample is known.
Specifically, denoting as $q^d_\mathtt{x}$ the distribution of activations for a given feature channel and domain $d$, an input $x^d\sim q^d_\mathtt{x}$ to the DA-layer can be normalized according to
\[
  \DA(x^d; \mu_d, \sigma_d) = \frac{x^d - \mu_d}{\sqrt{\sigma_d^2 + \epsilon}},
\]
where $\mu_d = \E_{x\sim q^d_\mathtt{x}}[x]$, $\sigma^2_d = \Var_{x\sim q^d_\mathtt{x}}[x]$ and $\epsilon>0$ is a small constant to avoid numerical issues.
When the statistics $\mu_d$ and $\sigma^2_d$ are computed over the current batch, this equates in practice to applying standard Batch Normalization separately to the samples of each domain.

As mentioned above, this approach requires full domain knowledge, as, for each $d$, $\mu_d$ and $\sigma^2_d$ need to be calculated on the specific samples belonging to $d$.
In our case, however, while the target is clearly distinct from the source, we do not know which specific source domain most or even all of the source samples belong to.
To tackle this issue, we propose to model the layer's input distribution as a mixture of Gaussians, with one component for each domain.
Specifically, we define a global input distribution $q_\mathtt{x} = \sum_d \pi_d q^d_\mathtt{x}$, where $\pi_d$ is the probability of sampling from domain $d$, and $q^d_\mathtt{x} = \mathcal{N}(\mu_d, \sigma^2_d)$ is the domain-specific distribution for $d$: a normal with mean $\mu_d$ and variance $\sigma_d^2$.
Given a batch of samples $\set{B}=\{x_i\}_{i=1}^\con{b}$, a maximum likelihood estimate of the parameters $\mu_d$ and $\sigma_d^2$ is given by
\begin{equation}
\label{eqn:mixture-params}
\begin{aligned}
  \mu_d &= \sum_{i=1}^\con{b} \alpha_{i,d} x_i, &
  \sigma_d^2 &= \sum_{i=1}^\con{b} \alpha_{i,d} (x_i - \mu_d)^2,
\end{aligned}
\end{equation}
where
\begin{equation}
\label{eqn:weights}
  \alpha_{i,d} = \frac{q_\mathtt{d|x}(d\mid x_i)}{\sum_{i=1}^\con{b} q_\mathtt{d|x}(d\mid x_i)},
\end{equation}
and $q_\mathtt{d|x}(d\mid x_i)$ is the conditional probability of $x_i$ belonging to $d$, given $x_i$.
Clearly, the value of $q_\mathtt{d|x}$ is known for all samples for which we have domain information.
In all other cases, the missing domain assignment probabilities are inferred from data, using the \emph{domain prediction} network branch which will be detailed in Section~\ref{sec:domain-prediction}.
Thus, from the perspective of the alignment layer, these probabilities become an additional input, which we denote as $w_{i,d}$ for the predicted probability of $x_i$ belonging to $d$.

By substituting $w_{i,d}$ for $q_\mathtt{d|x}(d\mid x_i)$ in \eqref{eqn:mixture-params} and \eqref{eqn:weights}, we obtain a new set of empirical estimates for the mixture parameters, which we denote as $\hat{\mu}_d$ and $\hat{\sigma}^2_d$.
These parameters are used to normalize the layer's inputs according to
\begin{equation}
\label{eqn:normalization}
  \mDA(x_i, \vct{w}_i; \vct{\hat{\mu}}, \vct{\hat{\sigma}}) = \sum_{d\in \set D} w_{i,d} \frac{x_i - \hat{\mu}_d}{\sqrt{\hat{\sigma}_d^2 + \epsilon}},
\end{equation}
where $\vct{w}_i=\{w_{i,d}\}_{d\in\set D}$, $\vct{\hat{\mu}}=\{\hat{\mu}_d\}_{d\in\set D}$ and $\vct{\hat{\sigma}}=\{\hat{\sigma}^2_d\}_{d\in\set D}$.
As in previous works \cite{carlucci2017autodial,carlucci2017just,ioffe2015batch}, during back-propagation we calculate the derivatives through the statistics and weights, propagating the gradients to both the main input and the domain assignment probabilities.

\subsection{Domain prediction}
\label{sec:domain-prediction}

As explained in the previous Section~\ref{sec:dalayers}, our mDA-layers take as input a set of domain assignment probabilities for each input sample, which need to be predicted. 
While different mDA-layers in a network have in general different input distributions, the assignment of sample points to domains should be coherent across them. Specifically, sample points at different mDA-layers corresponding to a single input element to the network should share the same probabilities.
As a practical example, in the typical case in which mDA-layers are used in a CNN to normalize convolutional activations, the network would predict a single set of probabilities for each input image, which would then be given as input to all mDA-layers and broadcasted across all spatial locations and feature channels corresponding to that image.
Following these consideration, we compute domain assignment probabilities using a distinct section of the network, which we call the \emph{domain prediction} branch, while we refer to the main section of the network as the \emph{classification} branch.
The two branches share the bottom-most layers and parameters as depicted in Figure~\ref{fig:teaser}.
The domain prediction branch is implemented as a minimal set of layers followed by a soft max operation with $\con{k}$ outputs for the $\con{k}$ latent source domains (more details follow in Section~\ref{sec:experiments}).
As the domain membership of target samples is always assumed to be known, we do not predict domain assignment probabilities for the target.
Furthermore, for each sample $x_i$ with known domain membership $\hat d$, we fix in each mDA-layer $w_{i,d}=1$ if $d=\hat d$, otherwise $w_{i,d}=0$ .

We split the network into a domain prediction branch and classification branch at some low level layer.
This choice is motivated by the observation~\cite{aljundi2016lightweight} that features tend to become increasingly more domain invariant going deeper into the network, meaning that it becomes increasingly harder to compute a sample's domain as a function of deeper features.
In fact, as pointed out in~\cite{carlucci2017autodial}, this phenomenon is even more evident in networks that include Domain Alignment layers.

\subsection{Training the network}
\label{sec:loss}

We want to estimate $\theta\in \Theta$, which comprises all trainable parameters of the classification and domain prediction branches, while taking advantage of both labeled and unlabeled data.
A main difficulty lies in the fact that, when employing a discriminative model, the unlabeled samples cannot be used to express the data likelihood.
However, following the approach sketched in~\cite{carlucci2017autodial}, we can exploit the unlabeled data to define a prior distribution over the network's parameters.
By doing this, we define a posterior distribution over $\theta$ given all data and labels as follows:
\begin{multline}
\label{eqn:posterior}
  \pi(\theta|\set{S},\set{T},\hat{\set{S}}) \propto \pi(y_\set{S}|x_\set{S},\theta)\\
  \cdot\pi(d_{\hat S}|x_{\hat{\set S}},\theta)
  \pi(\theta|\set{T})
  \pi(\theta|x_\set{S\setminus\hat S}),
\end{multline}
where for notational convenience we have omitted some dependences.
By maximizing \eqref{eqn:posterior} over $\Theta$ we obtain a maximum-a-posterior estimate $\hat{\theta}$ for the parameters:
\begin{equation}
\label{eqn:map}
  \hat{\theta} \in \argmax_{\theta \in \Theta} \pi(\theta|\set{S},\set{T},\hat{\set{S}}).
\end{equation}

The first term on the right hand side of \eqref{eqn:posterior} is the likelihood of $\theta$ w.r.t. the source dataset, and can be written as
\begin{equation}
\label{eqn:likelihood-class}
  \pi(y_\set{S}|x_\set{S},\theta) = \prod_{i=1}^\con{n} f_C^\theta(y_i^s; x_i^s)
\end{equation}
due to the i.i.d. assumption on the training samples.
Here we denote by $f_C^\theta(y_i^s; x_i^s)$ the output of the \emph{classification branch} of the network for a source sample, \ie the predicted probability of $x_i^s$ having class $y_i^s$, and, for convenience of notation, we omit the dependence of $f_C^\theta$ on the target samples induced by the mDA-layers.
Similarly, the second term in \eqref{eqn:posterior} is the likelihood of $\theta$ w.r.t. the known domains:
\[
  \pi(d_{\hat {\set S}}|x_{\hat {\set{S}}},\theta) = \prod_{x_i \in x_{\hat {\set{S}}}} f_D^\theta(d_i; x_i),
\]
where $d_i$ is the domain corresponding to $x_i\in x_{\hat {\set S}}$.
In the previous equation, $f_D^\theta(d; x)$ denotes the output of the \emph{domain prediction} branch for a sample $x$ and domain $d$, \ie the predicted probability of $x$ belonging to $d$.

To define our prior $\pi(\theta|\set{T})$ over the parameters, we exploit all available unlabeled data, biasing our classifier towards exhibiting low uncertainty on the unlabeled samples, similarly to~\cite{carlucci2017autodial}. However, in addition, we introduce a prior term $\pi(\theta|x_{\set{S}\setminus\hat{\set S}})$, which exploits source sample points with missing domain labels.
Uncertainty when predicting class labels can be measured in terms of the empirical entropy
\begin{equation*}
  h_C(\theta|x_\set{S}) = -\frac{1}{\con{m}} \sum_{i=1}^\con{m} \sum_{y\in\set{Y}} f_C^\theta(y; x_i^t) \log f_C^\theta(y; x_i^t),
\end{equation*}
and similarly for the uncertainty when predicting domains.
\begin{equation*}
  h_D(\theta|x_{\set{S}\setminus\hat{\set S}}) = -\frac{1}{|x_{\set{S}\setminus\hat{\set S}}|} \sum_{x\in x_{\set{S}\setminus\hat{\set S}}} \sum_{i=1}^\con{k} f_D^\theta(s_i; x) \log f_D^\theta(s_i; x).
\end{equation*}
Now, $\pi(\theta|\set{T})$ can be obtained as the distribution with maximum entropy under the constraints $\int \pi(\theta|x_\set{S}) h_C(\theta|x_\set{S}) d\theta = \varepsilon_C$ and, similarly, $\pi(\theta|x_{\set{S}\setminus\hat{\set S}})$ can be regarded as a maximum entropy distribution under the constraint $\int \pi(\theta|x_{\set{S}\setminus\hat{\set S}}) h_D(\theta|x_{\set{S}\setminus\hat{\set S}}) d\theta = \varepsilon_D$, where $\varepsilon_C>0$ and $\varepsilon_D>0$ define the desired average uncertainties for class and domain predictions, respectively.
These optimization problems can be shown to have solutions:
\begin{equation*}
\begin{aligned}
  \pi(\theta|\set{T}) &\propto \exp(-\lambda_C h_C(\theta|\set{T}))\\
  \pi(\theta|x_{\set{S}\setminus\hat{\set S}}) &\propto \exp(-\lambda_D h_D(\theta|x_{\set{S}\setminus\hat{\set S}})),
\end{aligned}  
\end{equation*}
where $\lambda_C$ and $\lambda_D$ are the Lagrange multipliers corresponding to $\varepsilon_C$ and $\varepsilon_D$, respectively.

In practice, the optimization in \eqref{eqn:map} can be replaced by the equivalent minimization of the negative logarithm of the likelihood, obtaining our loss function:
\begin{equation}
\label{eqn:loss}
\begin{aligned}
  L(\theta) = &- \frac{1}{\con{n}} \sum_{i=1}^\con{n} \log f_C^\theta(y_i^s; x_i^s) \\
    &- \lambda_t \frac{1}{|x_{\hat {\set{S}}}|} \sum_{x_i \in x_{\hat {\set{S}}}} \log f_D^\theta(d_i; x_i) \\
    &- \lambda_C \frac{1}{\con{m}} \sum_{i=1}^\con{m} \sum_{y\in\set{Y}} f_C^\theta(y; x_i^t) \log f_C^\theta(y; x_i^t) \\
    &- \lambda_D \frac{1}{|x_{\set{S}\setminus\hat{\set S}}|} \sum_{x\in x_{\set{S}\setminus\hat{\set S}}} \sum_{i=1}^\con{k} f_D^\theta(s_i; x) \log f_D^\theta(s_i; x).
\end{aligned}
\end{equation}
The four terms, balanced by the hyper-parameters $\lambda_t$, $\lambda_C$ and $\lambda_D$, can be interpreted as two log-losses and two entropy losses applied to the classification and domain prediction branches of the network, respectively to samples with known and unknown labels.
Interestingly, since the classification branch has a dependence on the domain prediction branch via the mDA-layers, by optimizing \eqref{eqn:loss}, the network learns to predict domain assignment probabilities that result in a low classification loss.
In other words, the network is free to predict domain memberships that do not necessarily reflect the real ones, as long as this helps improving its classification performance.


\section{Experiments}
\label{sec:experiments}
\subsection{Experimental Setup}
\paragraph{Datasets.} In our evaluation we consider several common DA benchmarks: the combination of the USPS~\cite{friedman2001elements}, MNIST~\cite{lecun1998gradient} and MNIST-m~\cite{ganin2014unsupervised} datasets, the Office-31 \cite{saenko2010adapting} dataset, Office-Caltech \cite{gong2012geodesic} and the 
PACS \cite{li2017deeper} dataset.

\textbf{MNIST, MNIST-m and USPS} are three standard datasets for digits recognition. 
USPS \cite{friedman2001elements} is a dataset built using digits scanned from U.S. envelopes, MNIST~\cite{lecun1998gradient} is the popular benchmark for digits recognition and MNIST-m~\cite{ganin2014unsupervised} its counterpart obtained by blending the original images with colored patches extracted from BSD500 photos~\cite{arbelaez2011contour}. 
Due to their different representations (\eg colored vs gray-scale), these datasets have been adopted as a DA benchmark by many previous works \cite{ganin2014unsupervised,bousmalis2016domain,bousmalis2016unsupervised}. Here, we consider a multi source DA setting, using MNIST and MNIST-m as sources and USPS as target, training on the union of the training sets and testing on the test set of USPS.

\textbf{Office-31} is a standard DA benchmark which contains images of 31 object categories collected from 3 different sources: Webcam (W), DSLR camera (D) and the Amazon website (A). Following \cite{xiong2014latent}, we perform our tests in the multi-source setting, where each domain is in turn considered as target, while the others are used as source.

\textbf{Office-Caltech} \cite{gong2012geodesic} is obtained by selecting the subset of $10$ common categories in the Office31 and the Caltech256 \cite{griffin2007caltech} datasets.
It contains $2533$ images, about half of which belong to Caltech256. The different domains are Amazon (A), DSLR (D), Webcam (W) and Caltech256 (C).
In our experiments we consider the set of source\slash{}target combinations used in~\cite{gong2013reshaping}.

\textbf{PACS} \cite{li2017deeper} is a recently proposed benchmark which is especially interesting due to the significant domain shift between different domains. It contains images of 7 categories (\textit{dog, elephant, giraffe, guitar, horse}) extracted from 4 different representations, \ie Photo (P), Art paintings (A), Cartoon (C) and Sketch (S). Following the experimental protocol in \cite{li2017deeper}, we train our model considering 3 domains as sources and the remaining as target, using all the images of each domain.
Differently from \cite{li2017deeper} we consider a DA setting (\ie target data is 
available at training time) and we do not address the problem of domain generalization.

\myparagraph{Networks and training protocols.}
We apply our approach to three different CNN architectures: the MNIST network described in \cite{ganin2014unsupervised}, AlexNet~\cite{krizhevsky2012imagenet} and ResNet~\cite{he2016deep}. We choose AlexNet due to its widespread use in state of the art DA approaches~\cite{ganin2014unsupervised,carlucci2017autodial,long2015learning,long2016unsupervised}, while ResNet is taken as an exemplar for recent architectures employing batch-normalization layers. Both AlexNet and ResNet are first pre-trained on ImageNet and then fine-tuned on the datasets of interest. The MNIST architecture in \cite{ganin2014unsupervised,ganin2016domain} is chosen following previous works considering digits datasets.

For the evaluation on digits datasets we employ the MNIST architecture described in \cite{ganin2014unsupervised}. 
Since the original architecture does not contain BN layers, we add mDA-layers after each layer with parameters. We train the architecture following the schedule defined in \cite{ganin2014unsupervised}, with a batch-size containing 128 images per domain.  
{The side-branch starts from the \texttt{conv1} layer, applies a second convolution with the same parameters of \texttt{conv2} and a fully-connected layer with 100 output channel
, before the final domain-classifier.} 

For the experiments on the Office-31 and Office-Caltech datasets we employ the AlexNet architecture. We follow a setup similar to the one proposed in~\cite{carlucci2017autodial,carlucci2017just}, fixing the parameters of all convolutional layers with mDA-layers inserted following each fully-connected layer and before their corresponding activation functions. 
The domain prediction branch 
is attached to the last pooling layer following \texttt{conv5}. It is composed of a global average pooling, followed by a fully connected layer and a softmax operation to produce the final domain probabilities. The training schedule and hyperparameters are set following \cite{carlucci2017autodial}.

For the experiments on the PACS dataset we consider the ResNet architecture and we choose the 18-layers setup described in~\cite{he2016deep} and denoted as ResNet18.
This architecture comprises an initial $7\times 7$ convolution, denoted as \texttt{conv1}, followed by 4 main modules, denoted as \texttt{conv2} -- \texttt{conv5}, each containing two residual blocks. To apply our approach, we replace each Batch Normalization layer in the network with an mDA-layer.
The domain prediction branch is attached to \texttt{conv1}, and is formed by adding a residual block (with the same number of filters as the ones in \texttt{conv2}) and a global average pooling layer followed by a fully connected layer 
and a softmax.
{For training we use a weight-decay of $10^{-6}$, with the same initial learning rate and momentum adopted for AlexNet.} The network is trained for 1200 iterations with a batch-size of 48, equally divided between the domains. The learning rate is scaled by a factor 0.1 after 75\% of the iterations. More details about the training procedures can be found in the supplementary material.

Regarding the hyper-parameters of our method,  we set the number of source domains $\con{k}$ equal to $Q-1$, where $Q$ is the number of different datasets used in each single experiment.
Following \cite{carlucci2017autodial}, in the experiments with AlexNet architecture we fix $\lambda_C=\lambda_D=0.2$. Similarly, for the experiments on digits classification, we keep the weights $\lambda_C$, $\lambda_D$ of the two entropy losses fixed to the same value (0.1). For ResNet we select the values $\lambda_C=0.1$ and $\lambda_D=0.0001$ through cross-validation, following the procedure adopted in \cite{long2013transfer,carlucci2017autodial}. When domain labels are available for a subset of source samples, we fix $\lambda_t=0.5$.


We implement 
all the models with the Caffe~\cite{jia2014caffe} framework and our evaluation is performed using a NVIDIA GeForce 1070 GTX GPU.  We initialize both AlexNet and ResNet networks through their models pre-trained on ImageNet. 
For AlexNet we take the pre-trained model available in Caffe, while for ResNet we use the converted version of the original model developed in Torch \footnote{\scriptsize\url{ https://github.com/HolmesShuan/ResNet-18-Caffemodel-on-ImageNet}}.

\subsection{Results}

In this section we report the results of our evaluation. We first analyze the proposed approach, demonstrating the advantages of considering multiple sources and discovering latent domains. We then compare the proposed method with state-of-the-art approaches. 
For all the experiments we report the results in terms of accuracy, repeating the experiments 5 times and averaging the results. 

\myparagraph{Analysis of the Proposed Approach.}
In a first series of experiments, we test the performance of our approach on the MNIST-MNIST-m to USPS benchmark.
We compare our method with different baselines: (i) the network trained on the union of all source domains (\textit{Single source (unified)}), (ii) the model which leads to the best performance among those trained on each single source domain (\textit{Best single source}) (iii) the domain adaptation method DIAL in \cite{carlucci2017just} which uses as source set the union of all source domains (\textit{DIAL \cite{carlucci2017just} - Single source (unified)}) and (iv) the DIAL model which leads to the best performance among those trained on each single source domain (\textit{DIAL \cite{carlucci2017just} - Best single source}). Moreover, we report the results of our approach in the ideal case where the multiple source domains are known and we do not need to discover them (\textit{Multi-source DA}).
For our approach, we consider several different values for the hyper-parameter $\con{k}$, \ie the number of discovered source domains.
All these methods are based on the MNIST network in \cite{ganin2014unsupervised} with the addition of BN layers.

Table~\ref{tab:digits} shows the results of our comparison. By looking at the table several observations can be made. First, there is a large performance gap between models trained only on source data and DA methods, confirming the fact that deep architectures do not solve the domain shift problem \cite{donahue2014decaf}. Second, in analogy with previous works on DA \cite{mansour2009domain,duan2009domain,sun2011two}, we found that considering multiple sources is beneficial for reducing the domain shift with respect to learning a model on the unified source set. Finally, and more importantly, when the domain labels are not available, our approach is successful in discovering latent domains and in exploiting this information for improving accuracy classification on target data, partially filling the performance gap between the single source models and Multi-source DA. Interestingly, the performance of the algorithm are comparable when the number of latent domains $\con{k}$ changes, highlighting the robustness of our model to different values of $\con{k}$.
This motivates our choice to always fix $\con{k}$ to the known number of domains in the next experiments.


\begin{table}[t]
			\caption{Digits datasets: comparison of different models in the multi-source scenario. MNIST (M) and MNIST-m (Mm) are taken as source domains, USPS (U) as target.\vspace{-5pt}} 
		\centering
		\scalebox{.9}{
		\begin{tabular}{ l | c  } 
			\hline
			Method & M-Mm to U\\
            	\hline
                Single source (unified) &  57.1\\
                Best single source & 59.8 \\
			DIAL \cite{carlucci2017just} - Single source (unified) &81.7 
            \\
          {DIAL \cite{carlucci2017just} - Best single source} & 81.9 \\
Ours $\con{k}=2$& 	82.5\\
Ours $\con{k}=3$& 	82.2\\
Ours $\con{k}=4$& 	82.7\\
Ours $\con{k}=5$& 82.4\\\hline\hline
{Multi-source DA}&84.2 
\\ \hline 
		\end{tabular}
        }
		\label{tab:digits}
        \vspace{-5pt}
\end{table}

\begin{table}[t]
			\caption{PACS dataset: comparison of different methods using the ResNet architecture. The first row indicates the target domain, while all the others are considered as sources.\vspace{-5pt}} 
		\centering
		\scalebox{.9}{
		\begin{tabular}{ l | c  c c  c | c } 
			\hline
			Method & Sketch & Photo & Art & Cartoon & Mean\\
            	\hline
                ResNet \cite{he2016deep} &60.1&92.9&74.7&72.4&75.0\\
DIAL \cite{carlucci2017just} &66.8&\textbf{97.0}&87.3&85.5&84.2\\
Ours &\textbf{69.6}&\textbf{97.0}&\textbf{87.7}&\textbf{86.9}&\textbf{85.3}\\\hline\hline
Multi-source DA & 71.6 & 96.6 & 87.5 & 87.0 & 85.7 \\ \hline 
		\end{tabular}
        }
		\label{tab:pacs}
        \vspace{-15pt}
\end{table}

In a second series of experiments we consider the PACS dataset. We compare the proposed approach with the original ResNet architecture trained only on source data and with DA method DIAL \cite{carlucci2017just} trained on the unified source set. As in the previous experiments, we report the results of the ideal multi-source DA setting, \ie our approach is applied to multiple {known} source domains.
Table \ref{tab:pacs} shows our results. As expected, DA models are especially beneficial when considering the PACS dataset. Moreover, the multi-source DA network outperforms the single source one.  
Remarkably, our model is able to infer domain information automatically without supervision. In fact, its accuracy is either comparable with the multi-source model (\ie for Photo, Art and Cartoon) or in between the single-source, \ie DIAL, and the multi-source models (\ie Sketch). 

Looking at the partial results, it is especially interesting to see that the improvements of our approach and the multi-source model over DIAL trained on the unified source set are especially significant when either the Sketch or the Cartoon domains are employed as target set. Since these domains are less represented in the ImageNet database, we believe that the corresponding features derived from the pretrained model are less discriminative. In this case DA methods play a significant role.

We also conduct experiments on the Office31 dataset. As baselines we consider the standard AlexNet architecture trained on source data, AlexNet with Batch Normalization added after each fully-connected layer and the DA model of \cite{carlucci2017just} with all 
source domains unified in a single set. Again, the multi-source DA model obtained assuming the domain labels known for each source sample is taken as upper bound.
The results reported in Table \ref{tab:ablation-office} trigger two main observations. First, in this dataset there is a small margin for improvement when using a multi-source model with respect to adopting a single source one. This is in accordance with findings in \cite{li2017deeper}, where it is shown that, with respect to PACS dataset, in Office31 the domain shift with deep features is limited and it is linked mainly to changes in background (\ie Webcam-Amazon, DSLR-Amazon) or acquisition camera (DSLR-Webcam). Second, in this case our approach only slightly improves performance over the single-source DA model, suggesting that 
accuracy in automatically inferring latent domains may not be sufficient for learning better target classifiers. 

\begin{figure}[t]
  \centering
  \includegraphics[width=0.98\columnwidth]{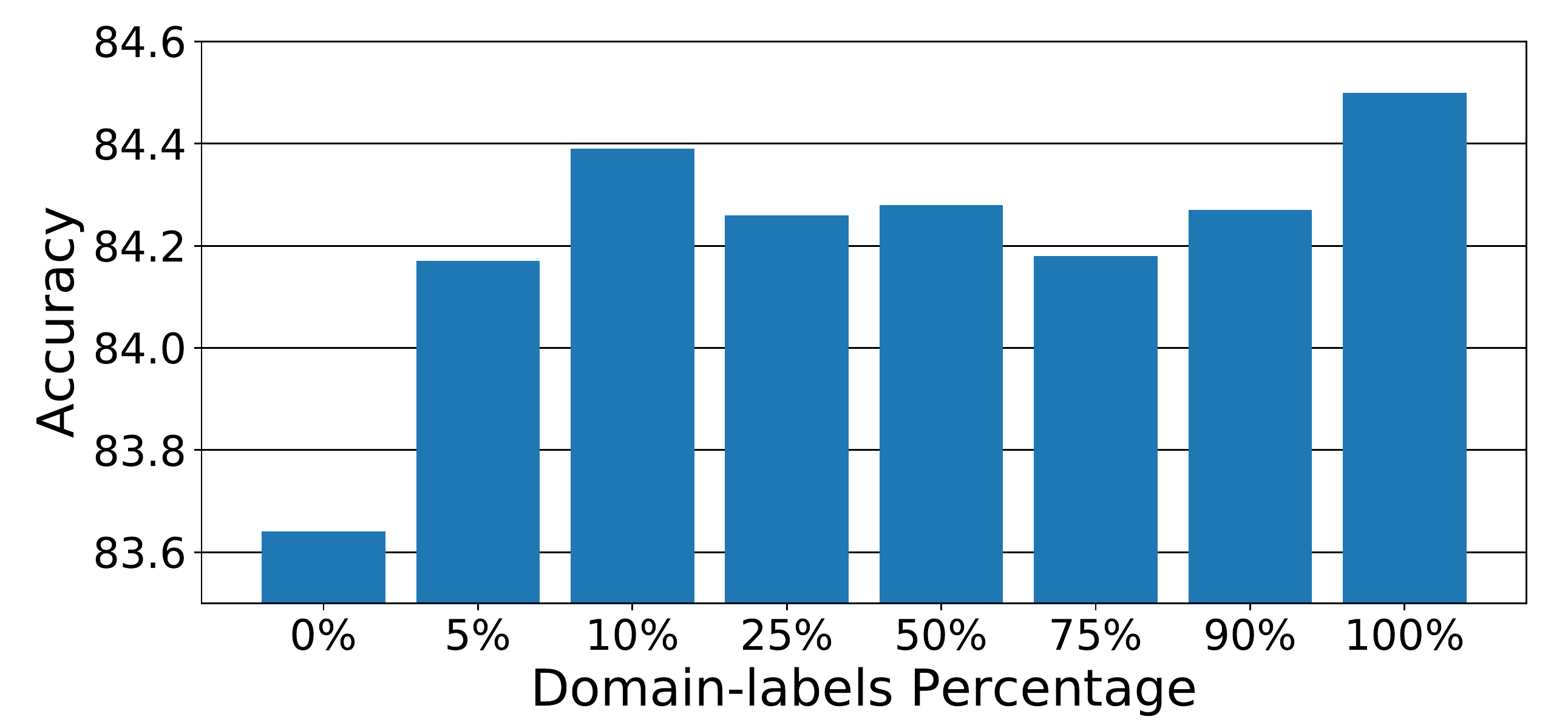}
  \vspace{-10pt}
  \caption{Office31 dataset. Performance at varying number of domain labels ($\%$) for source samples. 
  }\vspace{-15pt}
  \label{fig:office-bars}
\end{figure}

To further analyze this fact and to demonstrate the flexibility of our framework, we also perform an experiment in a semi-supervised setting. In particular, we consider different levels of supervision in terms of domain information and analyze how the performance of our method change at varying number of labeled source samples. 
The results of this experiment are reported in Fig. \ref{fig:office-bars}. Looking at the figure we can see that by using just few domain labels (5\% of the source samples), our model is able to completely fill the performance gap between the unsupervised and the multi-source model.
Furthermore, by increasing the level of supervision the accuracy saturates towards the value corresponding to the multi-source model.


\begin{table}[t]
			\caption{Office-31 dataset: comparison of different methods using AlexNet. In the first row we indicate the source (top) and the target domains (bottom).\vspace{-5pt}} 
		\centering
		\scalebox{.9}{
		\begin{tabular}{ l  r | c | c | c | c  } 
			\hline
			\multirow{2}{*}{Method}& Source& A-D & A-W & W-D  & \multirow{2}{*}{Mean}\\
           & Target & W & D & A  & \\\hline
           AlexNet \cite{krizhevsky2012imagenet}&		&89.1	&94.6&49.1	&77.6\\
AlexNet+BN &	&92.9&\textbf{95.2}	&60.1	&82.7\\

DIAL \cite{carlucci2017just} &
&94.3&93.8	&62.5	&83.5\\
Ours &
	&\textbf{94.6}&93.7	&\textbf{62.6}	&\textbf{83.6}\\






            \hline\hline
           Multi-source DA	&	&95.8&94.8	&62.9	&84.5\\ \hline
		\end{tabular}
        }
        \label{tab:ablation-office}
\end{table}

\myparagraph{Comparison with state of the art.}
In this section we compare the performance of our model with previous works on DA which also consider the problem of inferring latent domains \cite{hoffman2012discovering,xiong2014latent,gong2013reshaping}. As stated in Section \ref{sec:related}, there are no previous works adopting deep learning models (i) in a multi-source setting and (ii) discovering hidden domains. Therefore, the considered baseline methods \cite{hoffman2012discovering,xiong2014latent,gong2013reshaping} only employ handcrafted features. For these approaches we report results taken from the original papers. {To further analyze the impact of different feature representations, we also report results obtained running the method of Gong \etal \cite{gong2013reshaping} using features from the last layer of the AlexNet architecture. 
For a fair comparison, in this series of experiments we extract features from the \texttt{fc7} layer, without fine-tuning, applying mDA layers to these features and after the classifier.} 

We first consider the Office31 dataset, as this benchmark has been used in \cite{hoffman2012discovering,xiong2014latent}. Table \ref{tab:sota-office} shows the results of our comparison. Our model outperforms all the baselines, with a clear margin in terms of accuracy. Importantly, even when the method in \cite{gong2013reshaping} is applied to features derived from AlexNet, still our approach leads to higher accuracy. 
For the sake of completeness, in the same table we also report results from previous multi-source DA methods \cite{gopalan2014unsupervised,nguyen2015dash,lin2017cross}. Notice that also these methods are based on shallow models. While these approaches significantly outperform \cite{hoffman2012discovering} and \cite{xiong2014latent}, still their accuracy is much lower than ours.

\begin{table}[t]
			\caption{Office-31: comparison with state-of-the-art algorithms. In the first row we indicate the source (top) and the target domains (bottom).\vspace{-5pt}} 
		\centering
		\scalebox{.9}{
		\begin{tabular}{ l@{\hspace{-6ex}} r | c | c | c | c } 
			\hline
			 \multirow{2}{*}{Method}& Sources &A-D & A-W & W-D  & \multirow{2}{*}{Mean}\\          
             &Target & W & D & A  & \\  \hline
            Hoffman \etal \cite{hoffman2012discovering}&&24.8	&42.7	&12.8	&26.8\\
            Xiong \etal \cite{xiong2014latent}&&29.3	&43.6	&13.3	&28.7\\
            \hline
            Gong \etal (AlexNet) \cite{gong2013reshaping}	&&91.8&\textbf{94.6}	&48.9	&78.4\\
         Ours	&&\textbf{93.1}&
         94.3&\textbf{64.2}& \textbf{83.9}\\
            \hline\hline
                             Gopalan \etal  \cite{gopalan2014unsupervised} &&51.3	&36.1	&35.8	&41.1\\
          Nguyen \etal  \cite{nguyen2015dash} &&64.5	&68.6	&41.8	&58.3\\
Lin \etal \cite{lin2017cross} &&73.2	&81.3	&41.1	& 65.2\\ \hline
		\end{tabular}
        }
		\label{tab:sota-office}
\end{table}

To compare with \cite{hoffman2012discovering,gong2013reshaping}, we also consider the Office-Caltech dataset. {Following \cite{gong2013reshaping}, we test both single target (Amazon) and multi-target (Amazon-Caltech and Webcam-DSLR) scenarios, for our model can be easily extended to the latter case. We assume to know which samples belong to the source domains and which samples to the target domains. Then, we apply two different mDA modules: one for discovering latent source domains and one for discovering latent target domains. To this extent we need two domain prediction branches: in our implementation they share only the input features, while their parameters are independently learned. Notice that, since we do not assume to know the target domain to which a sample belongs, the task is even harder since we require a domain prediction step also at test time.} Again, our approach outperforms all baselines, even the method in \cite{gong2013reshaping} adopting features derived from AlexNet.

\begin{table}[t]
 			\caption{Office-Caltech dataset: comparison with state-of-the-art algorithms. In the first row we indicate the source (top) and the target domains (bottom).\vspace{-5pt} 
            } 
 		\centering
 		\scalebox{.8}{
 		\begin{tabular}{ l@{\hspace{-6ex}} r | c | c | c | c } 
        \hline
        		\multirow{2}{*}{Method}	 & Source& A-C & W-D & C-W-D  & \multirow{2}{*}{Mean}\\          
            & Target& W-D & A-C & A  & \\
 			\hline
                Gong \etal \cite{gong2013reshaping} - original& &41.7	&35.8	&41.0	&39.5\\
                Hoffman \etal \cite{hoffman2012discovering} - ensemble	&&31.7	&34.4	&38.9	& 35.0\\
               Hoffman \etal \cite{hoffman2012discovering} - matching &&39.6	&34.0	&34.6	&36.1\\
               Gong \etal \cite{gong2013reshaping} - ensemble &&38.7	&35.8	&42.8	&39.1\\
                Gong \etal \cite{gong2013reshaping}	- matching &&42.6	&35.5	&44.6	&40.9\\
             \hline
           Gong \etal (AlexNet) \cite{gong2013reshaping} &&{87.8}	&87.9	&93.6	&89.8\\
{Ours}&&\textbf{93.5}&	\textbf{88.2}&\textbf{93.7}&\textbf{91.8}\\
             \hline
 		\end{tabular}
         }
 		\label{tab:sota-office-caltech}
        \vspace{-10pt}
 \end{table}

%
%

\vspace{-3pt}
\section{Conclusions}
In this work we presented a novel deep DA model for automatically discovering latent domains within visual datasets. 
The proposed deep architecture is based on a side-branch which computes the assignment of a source sample to a latent domain. These assignments are then exploited within the main network by novel domain alignment layers which reduce the domain shift by aligning the feature distributions of the discovered sources and the target domains.
Our experimental results demonstrate the ability of our model to efficiently exploit the discovered latent domains for addressing challenging domain adaptation tasks.

{\myparagraph{Acknowledgements.} 
We acknowledge financial support from ERC grant 637076 - RoboExNovo and project \textit{DIGIMAP}, grant 860375, funded by the Austrian Research Promotion Agency (FFG).}

{\small
\bibliographystyle{ieee}
\bibliography{root}
}

\end{document}